\def\year{2021}\relax
\documentclass[letterpaper]{article}
\usepackage{aaai21} 
\usepackage{times} 
\usepackage{helvet} 
\usepackage{courier} 
\usepackage[hyphens]{url} 
\usepackage{graphicx} 
\usepackage{graphicx} 
\usepackage{natbib}
\usepackage{caption}
\usepackage{amsmath}
\usepackage{amsfonts}
\usepackage{amssymb}
\usepackage{booktabs}
\usepackage{bm}
\usepackage{microtype}

\usepackage{amsmath,amsfonts,amssymb}
\usepackage{bm}
\usepackage{algpseudocode}
\usepackage[ruled,vlined]{algorithm2e}

\begin{document}
%
\SetKwInput{KwInput}{Input}
\SetKwInput{KwOutput}{Output}

\title{IsoBN: Fine-Tuning BERT with Isotropic Batch Normalization}

\author{
Wenxuan Zhou,\qquad
Bill Yuchen Lin,\qquad
Xiang Ren\\}
\affiliations {
Department of Computer Science, University of Southern California, Los Angeles, CA \\
\{zhouwenx, yuchen.lin, xiangren\}@usc.edu}
\date{}

\maketitle
\begin{abstract}
Fine-tuning pre-trained language models (PTLMs), such as BERT and its better variant RoBERTa, has been a common practice for advancing performance in natural language understanding (NLU) tasks.
Recent advance in representation learning shows that \textit{isotropic} (i.e., unit-variance and uncorrelated) embeddings can significantly improve performance on downstream tasks with faster convergence and better generalization.
The isotropy of the pre-trained embeddings in PTLMs, however, is relatively under-explored.
In this paper, we analyze the isotropy of the pre-trained [CLS] embeddings of PTLMs with straightforward visualization, and point out two major issues: high variance in their standard deviation, and high correlation between different dimensions.
We also propose a new network regularization method, isotropic batch normalization (IsoBN) to address the issues, towards learning more isotropic representations in fine-tuning by dynamically penalizing dominating principal components.
This simple yet effective fine-tuning method yields about 1.0 absolute increment on the average of seven NLU tasks.
\end{abstract}

\section{Introduction}
\label{sec:intro} 
Pre-trained language models (PTLMs), such as BERT~\cite{Devlin2019BERTPO} and RoBERTa~\cite{Liu2019RoBERTaAR}, have revolutionized the area of natural language understanding (NLU).
Fine-tuning PTLMs has advanced performance on many benchmark NLU datasets such as GLUE~\cite{Wang2018GLUEAM}.
The most common fine-tuning method is to continue training pre-trained model parameters together with a few additional task-specific layers.
The PTLMs and task-specific layers are usually connected by the embeddings of \texttt{[CLS]} tokens, which are regarded as sentence representations.

\begin{figure}[t]
  \centering
  \includegraphics[width=0.95\linewidth]{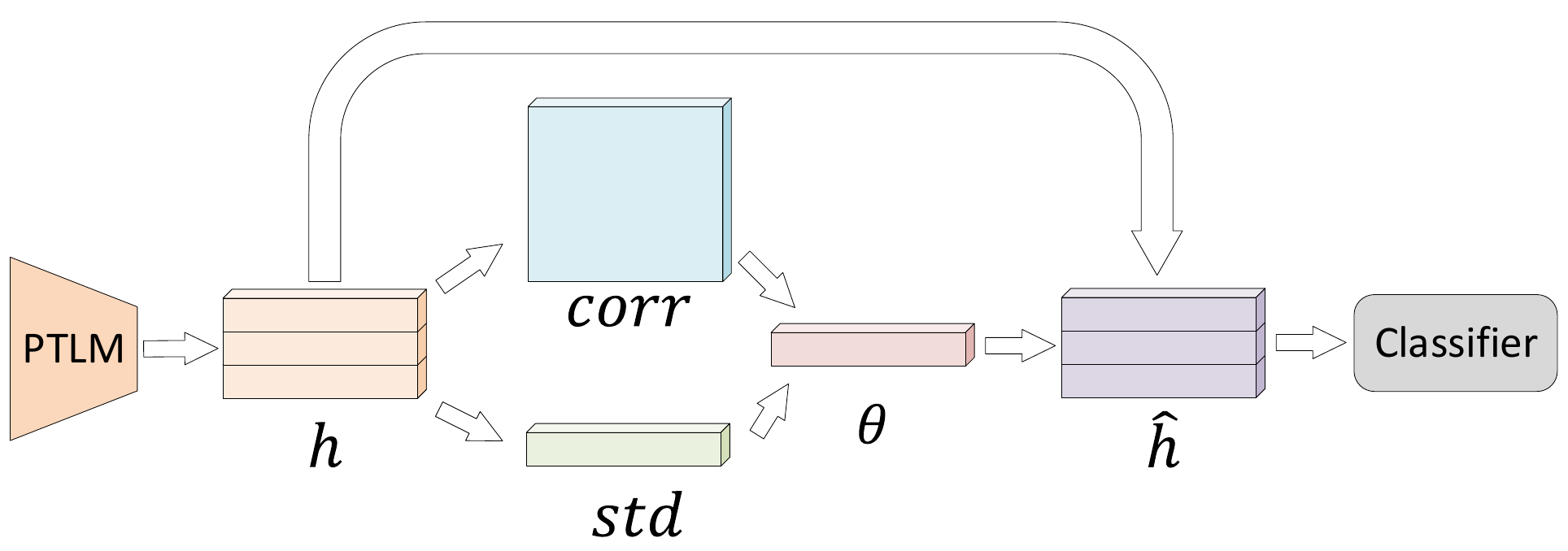}
  \caption{\textbf{Illustration of the isotropic batch normalization (IsoBN).} The \texttt{[CLS]} embedding is normalized by standard deviation and pairwise correlation coefficient to get a more isotropic representation.}
  \label{fig:illustration}
\end{figure}

Recent works on text representation~\cite{arora2016simple,mu2017allbutthetop,gao2019representation,Wang2020ImprovingNL} have shown that regularizing word embeddings to be more \textit{isotropic} (i.e., rotational invariant) can significantly improve their performance on downstream tasks.
An ideally isotropic embedding space has two major merits: a) all dimensions have the \textit{same variance} and b) all dimensions are \textit{uncorrelated} with each other.
These findings align with conventional \textit{feature normalization} techniques~\cite{cogswell2015reducing,ioffe2015batch,huang2018decorrelated}, which aim to transform input features into \textit{normalized\footnote{We use `normalized' to refer unit-variance in this paper.}, uncorrelated} representations for faster convergence and better generalization ability.

It, however, remains an open question that how isotropic the representations of PTLMs are.
Particularly, we want to understand the isotropy of pre-trained \texttt{[CLS]} embeddings in PTLMs, and how we can improve it towards better fine-tuning for downstream tasks.
In this paper, we first argue the reason why we want more isotropic embeddings for the \texttt{[CLS]} tokens (Section~\ref{sec:why_iso}).
Our analysis revels that the dominating principle components largely hinder the fine-tuning process to use knowledge in other components, due to the lack of isotropy.
Then, we analyze the isotropy of the pre-trained \texttt{[CLS]} embeddings.
There are two essential aspects of an isotropic embedding space: \textit{unit-variance} and \textit{uncorrelatedness}.
Thus, we start our analysis by visualizing the \textit{standard deviation} and Pearson \textit{correlation coefficient} of pre-trained \texttt{[CLS]} embeddings in BERT and RoBERTa on several NLU datasets.

Our visualization and quantitative analysis in Section~\ref{sec:analysis}
finds that: 1)  the \texttt{[CLS]} embeddings have very \textit{different variance}  (Sec.~\ref{ssec:std}); 2) the  \texttt{[CLS]} embeddings construct a few large clusters of dimensions that are \textit{highly correlated} with each other (Sec.~\ref{ssec:corr}).
Both findings indicate that pre-trained contextualized word embeddings are far from being isotropic, i.e., normalized and uncorrelated.
Therefore, these undesired prior bias from PTLMs may result in sub-optimal performance in fine-tuning for target tasks.

Given that pre-trained \texttt{[CLS]} embeddings are very anisotropic, a natural research question is then: 
\textbf{\textit{how can we regularize the fine-tuning process towards more isotropic embeddings?}}
There are two common methods for improving the isotropy of feature representations: whitening transformation and batch normalization~\cite{ioffe2015batch}.
However, both are not practically suitable in the scenario of fine-tuning PTLMs.
Whitening transformation  requires calculating the inverse of the covariance matrix, which are ill-conditioned in PTLMs' embeddings.
Unfortunately, calculating the inverse is thus numerically unstable, computationally expensive, and incompatible in half-precision training.
Batch normalization is proposed to alleviate the inverse-computation issue by assuming that the covariance matrix is \textit{diagonal}, which in turn completely ignores the influence of correlation between dimensions.

Motivated by the research question and limitations of existing works,
we propose a new network regularization method, isotropic batch normalization (IsoBN) in Section~\ref{sec:IsoBN}.
As shown in Fiure~\ref{fig:illustration}, the proposed method is based on our observation that the embedding dimensions can be seen as several groups of highly-correlated dimensions.
Our intuition is thus to assume that the absolute correlation coefficient matrix is a \textit{block-diagonal binary matrix}, instead of only a diagonal matrix.
The dimensions of the same group have an absolute correlation coefficient of 1 (duplicate of each other), and dimensions in different group of 0 (uncorrelated).

This method greatly reduces the computation efforts in calculating the inverse, and better models the characteristics of the pre-trained \texttt{[CLS]} embeddings.
Our experiments (Sec.~\ref{sec:exp}) show that the {IsoBN} indeed improves both BERT and RoBERTa in fine-tuning, yielding about 1.0 absolute increment on average of a wide range of 7 GLUE benchmark tasks.
We also empirically analyze the isotropy increment brought by IsoBN via explained variance, which clearly shows that IsoBN produces much more isotropic embeddings than conventional batch normalization method.
 
To the best of our knowledge, this work is the first one in studying the isotropy of the pre-trained \texttt{[CLS]} embeddings.
We believe our findings and the proposed IsoBN method will inspire interesting future research directions in improving pre-training language models as well as better fine-tuning towards more isotropy of PTLMs.

\section{Why isotropic \texttt{[CLS]} embeddings?}
\label{sec:why_iso}
We formally show our analysis on the principle components of the \texttt{[CLS]} embeddings.
Our findings reveal that with random weight initialization, the first few principle components usually take the majority of contributions for the prediction results.

\smallskip
\noindent
\textbf{Background knowledge.~}
 For text classification, the input text $x$ is first encoded by the PTLM to feature $\bm{h}$ (i.e., their [CLS] embeddings), and then classified by a random-initialized softmax layer:
\begin{align*}
    \bm{h} = \text{PTLM}(x) ~;~
    p_i = \frac{\exp(\bm{W}_i^T \bm{h})}{\sum_{j=1}^c \exp(\bm{W}_j^T \bm{h})},
\end{align*}
where $\bm{W}\in \mathcal{R}^{d \times c}$ is a random-initialized learnable parameter.
It learns towards mapping the underlying features extracted by PTLMs into target classes for input examples.

\begin{figure}[t]
  \centering
  \includegraphics[width=1\linewidth]{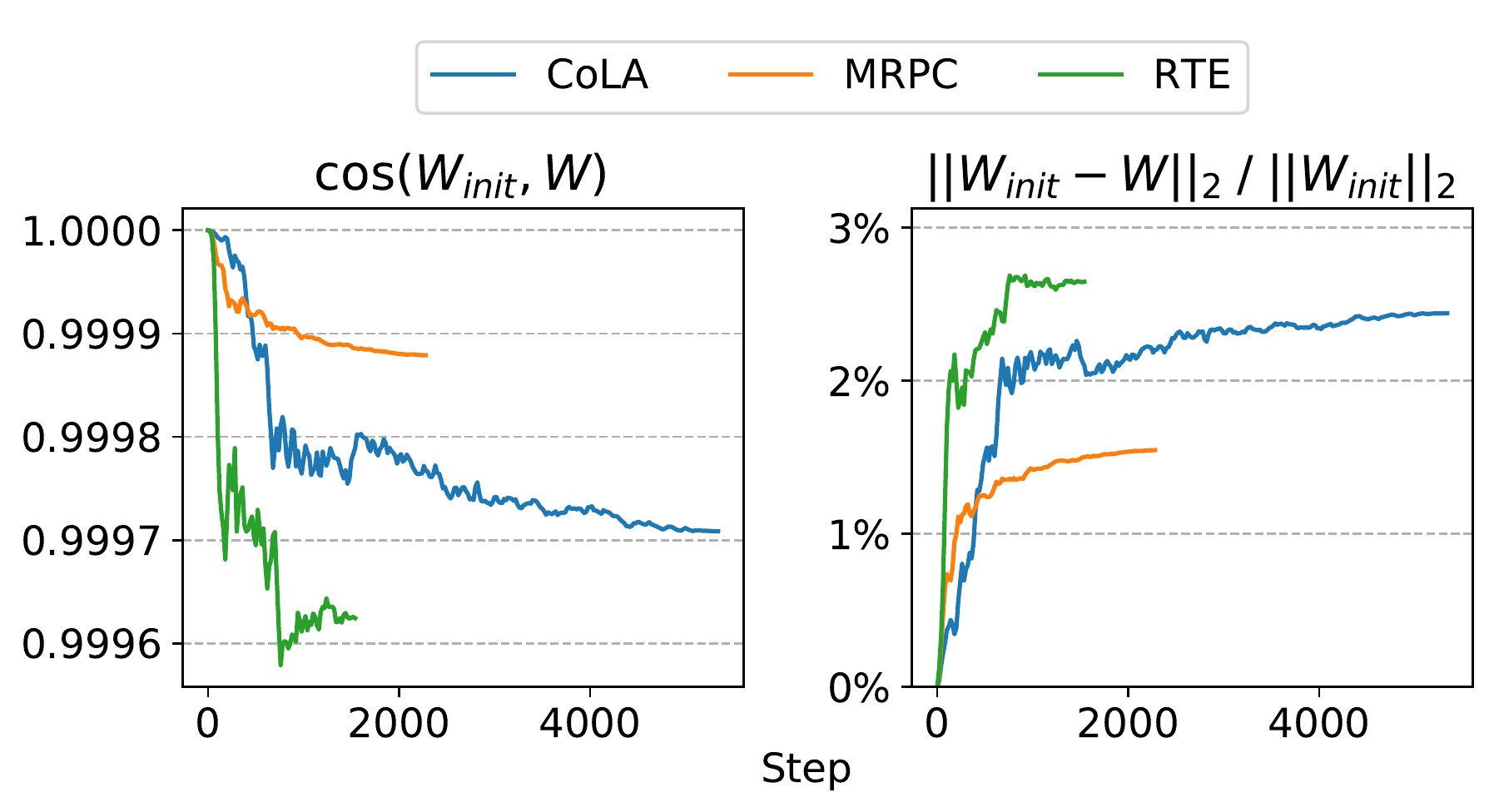}
  \caption{\textbf{Average cosine similarity and L2 distance of fine-tuned weight $\bm{W}$ to the initialized weight $\bm{W}_\text{init}$ during the entire training process.} Both measures suggest that the change of weight is very subtle.}
  \label{fig:difftoinit}
\end{figure}

Previous work~\cite{Dodge2020FineTuningPL} has shown that initialized weight $\bm{W}_\text{init}$ of classifier has a large impact on the model performance.
We further find that the final converged weights nearly remain the same to be the initialization after fine-tuning.
We visualize this surprising phenomenon in Figure~\ref{fig:difftoinit}.
We first project both $\bm{W}_\text{init}$ and $\bm{W}$ to the subspace spanned by the top 10 eigenvectors of $\text{Cov}\left(\bm{h}\right)$ to remove the unimportant components, then
use two similarity metrics (cosine similarity and L2 distance) to measure the difference between the initialized weight $\bm{W}_\text{init}$ and the fine-tuned weight $\bm{W}$.
We observe that the cosine similarity between $\bm{W}_\text{init}$ and $\bm{W}$ is extremely close to 1, and their L2 distance is close to 0.
It suggests that the weight $\bm{W}$ of the classifier in the fine-tuned model is almost determined in initialization.
For example, there is a 0.9997 cosine similarity between initialized and fine-tuned weights on COLA with RoBERTa.

\begin{figure}[t]
  \centering
  \includegraphics[width=1\linewidth]{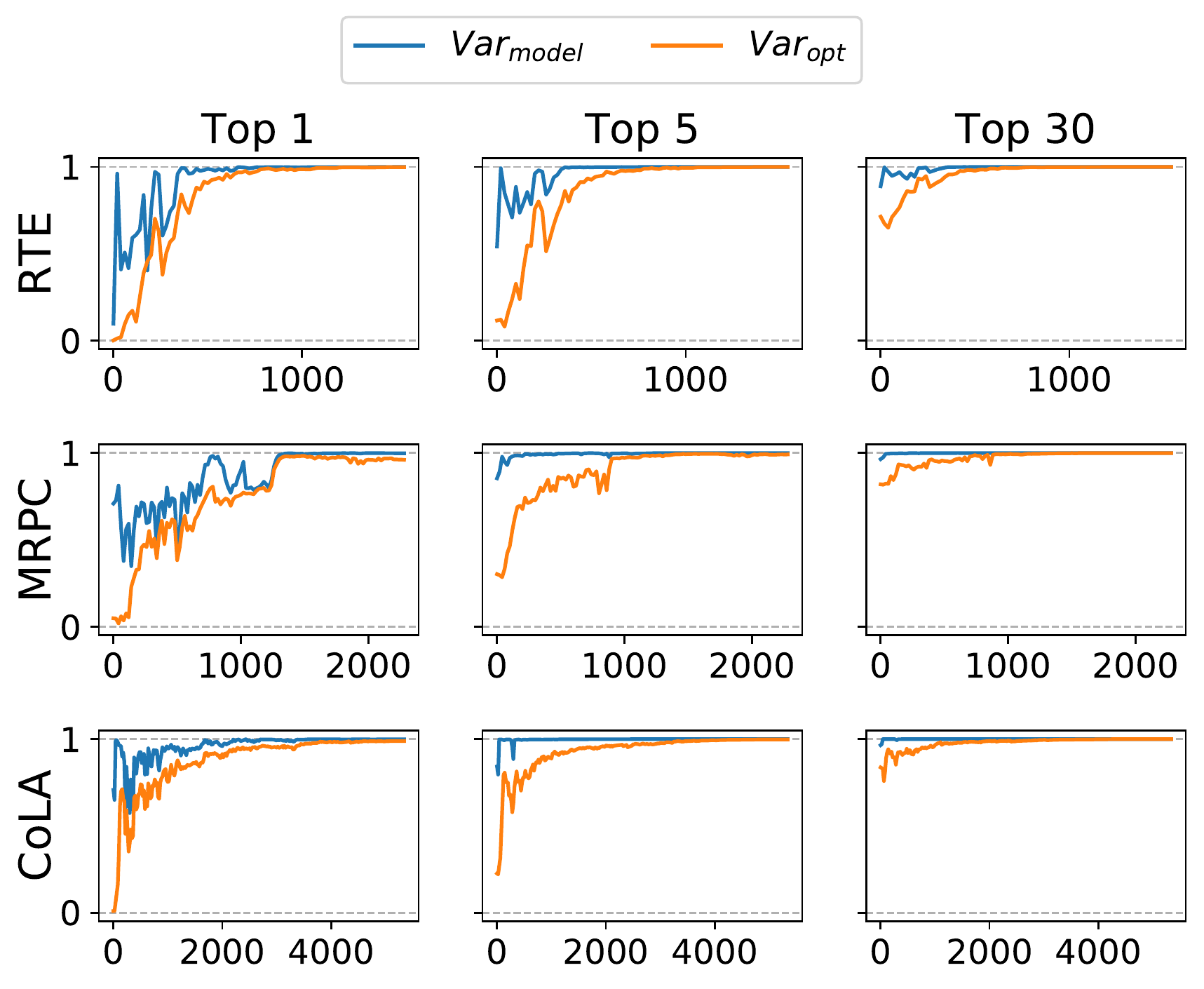}
  \caption{\textbf{The average percentage of variance of logit produced by the top 1, 5, and 30 principal components.} The top principal components are consistently exaggerated.}
  \label{fig:var}
\end{figure}

\smallskip
\noindent
\textbf{Dominating Principle Components.~} 
Knowing that the weight of classifier is nearly fixed, we infer that the classifier may not capture the discriminative information for classification during fine-tuning.
We measure the informativeness of each principal component, by comparing the variance of logits produced by it (between the fine-tuned classifier and the optimal classifier).
Specifically, we fit a logistic regression model on the entire training data using the scikit-learn~\cite{scikit-learn} framework to get the optimal classifier.
The variance $\text{Var}_i$ of logits by the $i^{th}$ principal component $(w_i, \bm{v}_i)$ is calculated by: $\text{Var}_i = w_i \cdot (\bm{W}^T \bm{v}_i)^2$,
where $w_i$, $\bm{v}_i$ are the $i^{th}$ eigenvalue and eigenvector.
Intuitively, a decent classifier should maximize the variance along informative principal components and minimize irrelevant ones.
We show the average proportion of variance in Figure~\ref{fig:var}.

As shown in our experiments (Sec.\ref{sec:exp}), the top principal components are constantly exaggerated by the fine-tuned weights, especially at the beginning of training process.
The first principal component accounts for over 90\% variance of logits throughout the training, which thus hinder learning from other useful components.
This motivates us to  penalize the top principal components dynamically during the fine-tuning for avoid losing knowledge of PTLMs due to dominating components.

\section{How isotropic are \texttt{[CLS]} embeddings?}
\label{sec:analysis}

The \texttt{[CLS]} embeddings of PTLMs, regarded as sentence representations, are directly used for fine-tuning (e.g., BERT and RoBERTa) towards a wide range of downstream tasks.
Given its impact in fine-tuning, we want to understand their isotropy.
As we know unit-variance and uncorrelatedness  are two essential aspects of an isotropy space,
we start our investigation by analyzing the \textit{standard deviation} and Pearson \textit{correlation coefficient}.

Specifically, we take the corpus of four popular NLU tasks (MRPC, RTE, COLA, and STS-b) from the GLUE benchmark datasets~\cite{wang2018glue} and then analyze their pre-trained \texttt{[CLS]} embeddings in terms of {standard deviation} (Section~\ref{ssec:std}) and {correlation} (Section~\ref{ssec:corr}) respectively.

\subsection{Analysis of Standard Deviation} 
\label{ssec:std}
To visualize the standard deviation of embeddings, we take the input sentences of the whole training dataset to PTLMs, then calculate the standard deviation on their \texttt{[CLS]} embeddings, and finally obtain the distribution.

The standard deviation of a nearly isotropic embedding space should concentrate on a very small range of values. 
Simply put,
an ideally isotropic embedding space should have a small variance of the distribution of their standard deviation, i.e.,  all dimensions of \texttt{[CLS]} embeddings should have almost the same standard deviation.

\begin{figure}[t]
  \centering
  \includegraphics[width=1\linewidth]{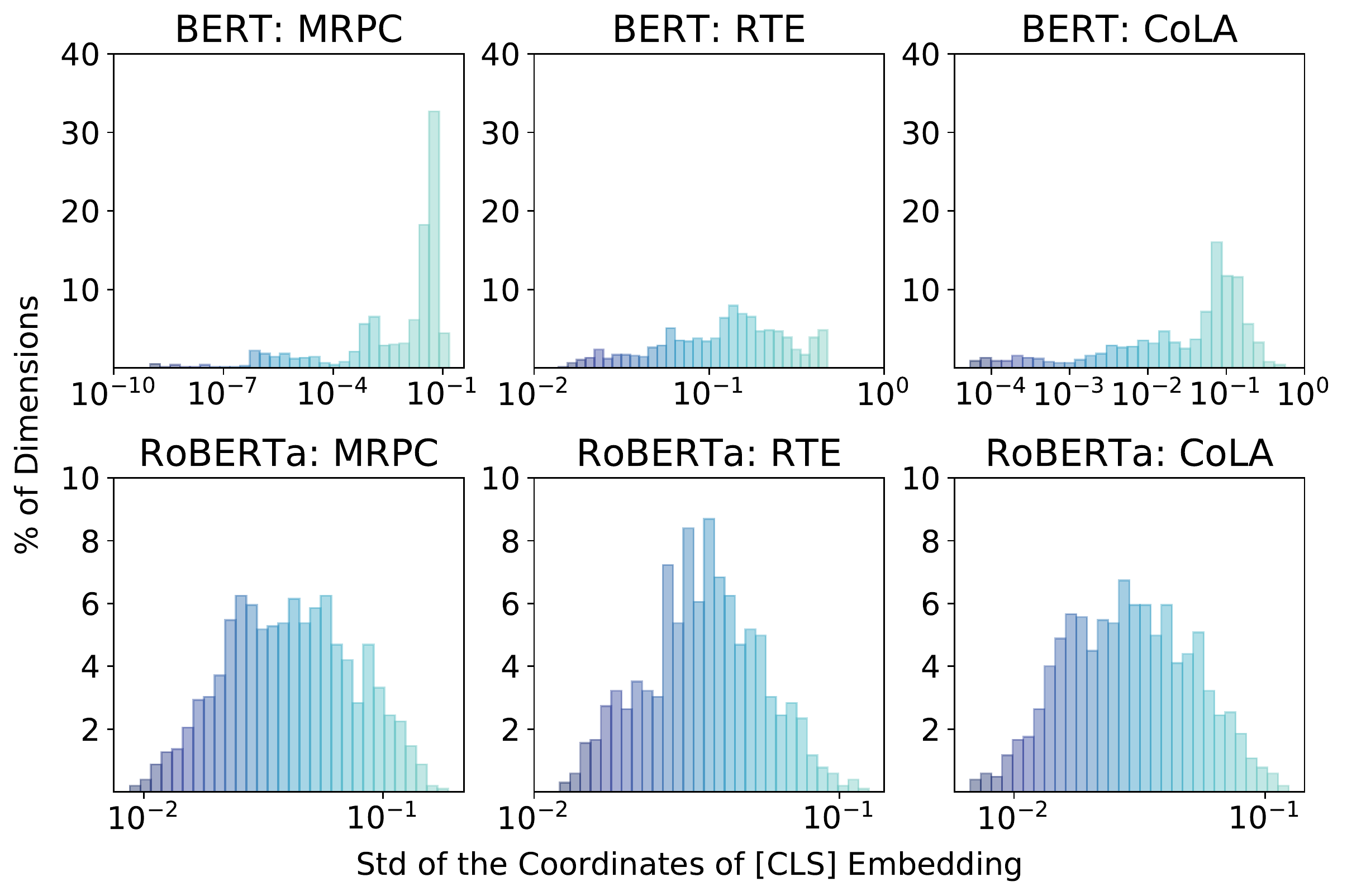}
  \caption{\textbf{The distribution of the standard deviation (std) of pre-trained \texttt{[CLS]}  embeddings.} We show the results of BERT-base and RoBERTa-Large on 4 NLU datasets. Note that an (nearly) isotropic embedding space should have (almost) zero variance on the std (i.e., 100\% dimensions have the same std).}
  \label{fig:std}
\end{figure}

As shown in Figure~\ref{fig:std},
we can see that both BERT and RoBERTa do not have such desired property for pre-trained \texttt{[CLS]} embeddings. 
The standard deviations of the embeddings vary in a very wide range of values (e.g.,  $[10^{-10}, 1]$ in BERT for MRPC).
Interestingly,
we can see that RoBERTa is evidently better than BERT from this perspective (e.g., usually ranging in $[0.01,1]$).
However, the \texttt{[CLS]}  embeddings of RoBERTa are still far from being isotropic, as there is no significantly dominant centering standard deviation value.

\subsection{Analysis of Correlation Coefficient}
\label{ssec:corr}
Correlation between different dimensions of \texttt{[CLS]}  embeddings is an essential aspect of isotropy.
Embeddings with low correlation between dimensions usually show better generalization on downstream applications~\cite{cogswell2015reducing}.
It, however, is relatively ignored by many neural network regularization methods, such as batch normalization~\cite{ioffe2015batch}.

\begin{figure}[t]
  \centering
  \includegraphics[width=1\linewidth]{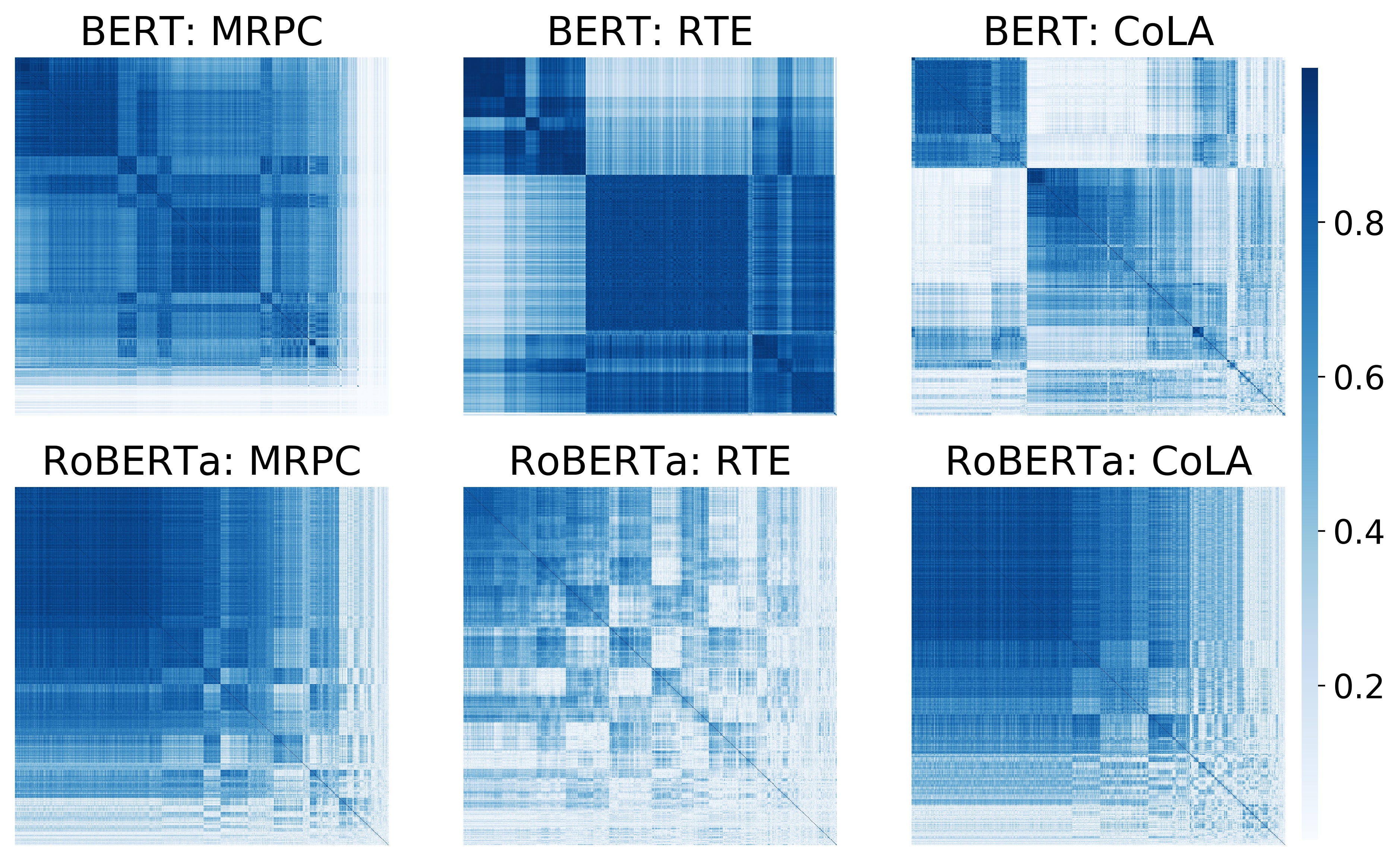}
  \caption{\textbf{Absolute Pearson correlation coefficients between dimensions of pre-trained \texttt{[CLS]} embeddings.}
  We show the results of BERT-base-cased and RoBERTa-Large on four NLU datasets. 
  Note that the dimension indexes of the matrices are re-arranged by the clustering results.
  Ideally, an isotropic embedding space should be 1 (darkest blue) on the diagonal and 0 on (white) other cells. 
  A dark block in a matrix means a cluster of features highly correlated with each other.
  }
  \label{fig:corr}
\end{figure}

In order to better visualize the Pearson correlation coefficient of \texttt{[CLS]} embeddings, 
we cluster the dimensions by their pairwise  coefficient, and then re-arrange the dimension index, such that highly correlated dimensions locate near each other.
The absolute value of correlations are shown in Figure~\ref{fig:corr}, where darker cells means higher correlation.

We can see that both BERT and RoBERTa usually have very high correlations between different dimensions (i.e., most cells are in dark blue), although the situation is less severe in a few cases such as BERT on CoLA and RoBERTa on RTE.
We find that BERT's embeddings have several large clusters of correlated features, while RoBERTa tends to have a single extreme large cluster.

In either case, such high correlation between embedding dimensions is harmful to future fine-tuning.
Recall that the \texttt{[CLS]} embeddings are usually connected to a linear classifier which is uniformly initialized.
In the beginning of the fine-tuning process,
the classifier will be biases to these features since they gain more importance in back-propagation.
This undesired prior prevents models to exploit other potentially value features, and thus require more training data or epochs to converge and generalize in downstream tasks.

We argue that these two findings together indicate that pre-trained language models are far from being isotropic (i.e., normalized and uncorrelated), and thus undesired prior bias may result in sub-optimal model performance for fine-tuning.

\section{Approach}
\label{sec:IsoBN}
Based on our analysis in Section~\ref{sec:analysis},
we propose a new regularization method, isotropic batch normalization towards learning more isotropic representations of the \texttt{[CLS]} tokens and thus better fine-tuning PTLMs.
We first introduce some background knowledge about \textit{whitening} and conventional \textit{batch normalization} methods (Section~\ref{ssec:bg}), then formally introduce the proposed IsoBN (Section~\ref{ssec:isobn}), and finally show the implementation details.

\subsection{Whitening and Batch Normalization}
\label{ssec:bg}

To improve the isotropy of feature representations,
there are two widely-used methods: 1) whitening transformation and 2) batch normalization~\cite{ioffe2015batch}.

\textbf{Whitening transformation} changes the input vector into a white noise vector, and can be defined as a transformation function as follows:
\begin{equation}
    \widehat{\bm{h}} = \Sigma^{-\frac{1}{2}} (\bm{h} - \bm{\mu} \cdot \bm{1}^T),
\end{equation}
where $\Sigma \in \mathcal{R}^{d \times d}$ is the covariance matrix of the input $\bm{h} \in \mathcal{R}^{d \times N}$, $\bm{\mu} \in \mathcal{R}^d$ is the mean of $\bm{h}$.
Thus, the transformation is a mapping from  $\mathcal{R}^{d \times N} \rightarrow \mathcal{R}^{d \times N}$.
This transformation produces a perfectly isotropic embedding space, 
where the dimensions are uncorrelated and have the same variance.
It can be applied in either  feature pre-processing~\cite{rosipal2001kernel} or neural network training~\cite{huang2018decorrelated}.
A similar method is to remove a few top principal components from the embedding space~\cite{arora2016simple,mu2017allbutthetop}.
However, these methods are hard to apply in fine-tuning PTLMs, as they require calculating the \textit{inverse of the covariance matrix}.
As shown in Section~\ref{ssec:corr}, the embeddings in PTLMs contains groups of highly-correlated dimensions.
Therefore,  the covariance matrices are ill-conditioned, and calculating the inverse is thus numerically unstable.
It is also computationally expensive and incompatible in half-precision training.

\textbf{Batch normalization} (BN) aims to simplify the inverse-computation problem by assuming that the covariance matrix is diagonal, thus the whitening function becomes:
\begin{equation}
    \widehat{\bm{h}} = \Lambda^{-1} (\bm{h} - \bm{\mu} \cdot \bm{1}^T),
\end{equation}
where $\Lambda=\text{diag}(\sigma_1, ..., \sigma_d)$ is a diagonal matrix consisting the standard deviation of each input dimension.
Batch normalization greatly improves the stability and model performance in training deep neural networks.

However, it completely ignores the influence of correlation in the embeddings, and thus not suitable for our interested \texttt{[CLS]} embeddings, where high correlation is a critical issue that needs to be addressed.
We seek to design a novel normalization method specially for fine-tuning PTLMs, which can be efficiently computed yet still improve representations towards isotropy property.

\subsection{Isotropic Batch Normalization}
\label{ssec:isobn}

Recall Figure~\ref{fig:corr}, from the correlation matrix of pre-trained embeddings, we observe that on most datasets, the correlation matrix is nearly \textit{block-diagonal}\footnote{A \textit{block diagonal} matrix is a block matrix that is a square matrix such that the main-diagonal blocks are square matrices and all off-diagonal blocks are zero matrices.}.
That is, the embedding dimensions form several clusters of highly-correlated dimensions.
Dimensions within the same cluster have an absolute correlation coefficient of nearly 1,
while dimensions from different clusters are almost uncorrelated.
\begin{figure}[t]
  \centering
  \includegraphics[width=1\linewidth]{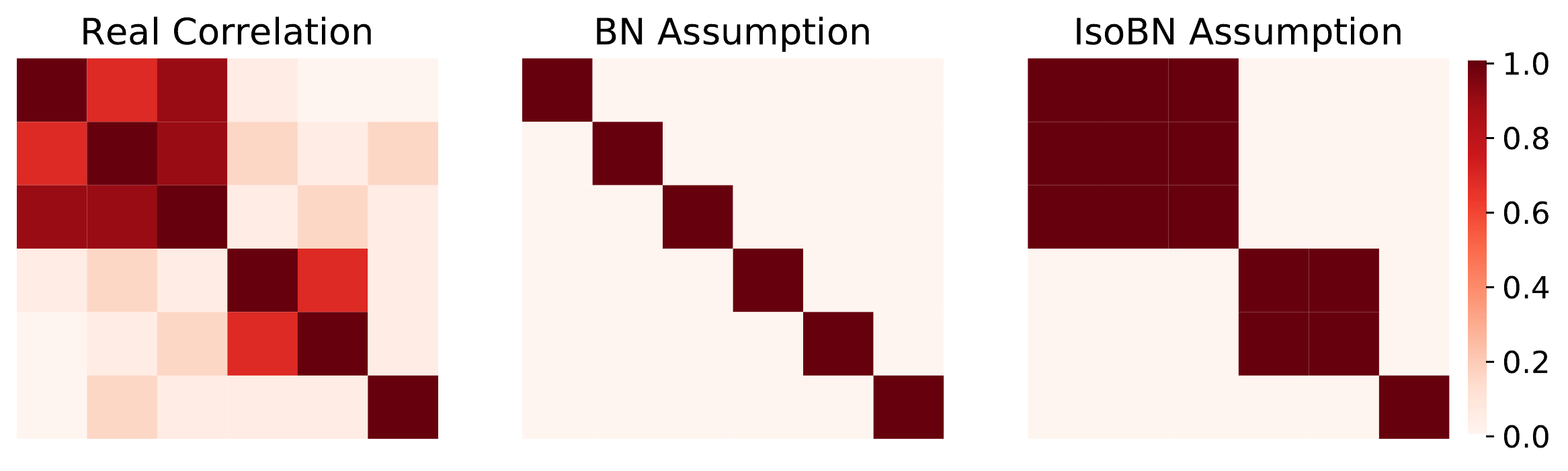}
  \caption{\textbf{Illustration of the assumption by batch normalization and our IsoBN with the reference of real correlation.} IsoBN assumes that the absolute correlation matrix is \textit{block-diagonal} while batch normalization completely ignores the correlation.}
  \label{fig:simp}
\end{figure}
Inspired by this, we propose an enhanced simplification of the covariance matrix.

We assume that the absolute correlation coefficient matrix is a \textbf{ block-diagonal binary matrix.} 
That is, the embedding dimensions can be clustered into $m$ groups $\mathcal{G}_1, ..., \mathcal{G}_m$, where dimensions of the same group have an absolute correlation coefficient of \texttt{1} (duplicate of each other), and dimensions in different group have a correlation coefficient of \texttt{0} (uncorrelated).
This assumption is illustrate in Figure~\ref{fig:simp} as a conceptual comparison.
Comparing with the conventional batch normalization, 
our assumption takes accounts of correlations and thus is an more accurate approximation of the realistic correlation coefficient matrices.
Thereby, instead of whitening the correlation matrix, we want the influence of each group of dimensions similar in the fine-tuning process.

We first normalize each dimension to unit-variance, similar to batch normalization, for convenience of further derivation.
This makes the dimensions in the same group exactly same to each other.
Then, for dimension $i\in \mathcal{G}_{g(i)}$, it is repeated in embeddings by $|\mathcal{G}_{g(i)}|$ times.
Therefore, the normalization transformation becomes:
\begin{equation}
    \widehat{\bm{h}}^{(i)} = \frac{1}{\sigma_i \cdot |\mathcal{G}_{g(i)}|} ({\bm{h}}^{(i)} - \mu_i \cdot \bm{1}^T).
\end{equation}

The dimensions of embeddings, however, are not naturally separable into hard group divisions.
Thus, we create a soft version of computing the size of a feature-group $|\mathcal{G}_{g(i)}|$ via the correlation coefficient matrix  $\rho$:
\begin{equation}
    |\mathcal{G}_{g(i)}| \xrightarrow{\sim} \gamma_i = \sum_{j=1}^d \rho_{ij}^2.
    \label{eq:gamma}
\end{equation}
This equation produces the same result as $|\mathcal{G}_{g(i)}|$ when our assumption holds in real correlation matrix.
Finally, our transformation can be written as:
\begin{equation}
    \widehat{\bm{h}}^{(i)} = \frac{1}{\sigma_i \cdot \gamma_i} ({\bm{h}}^{(i)} - \mu_i \cdot \bm{1}^T).
    \label{eq:our_norm}
\end{equation}

The major difference between our method and conventional batch normalization is the introduction of the $\gamma$ term, as a way to explicitly consider correlation between feature dimensions.
As shown in our experiments (Section~\ref{sec:exp}), $\gamma$ can greatly improve the isotropy of embedding.
We name our proposed normalization method as \textit{isotropic batch normalization} ({IsoBN}), as it is towards more isotropic representations during fine-tuning.

The {IsoBN} is applied right before the final classifier.
In experiments, we find that a modified version of IsoBN achieves better performance.
The mean $\bm{\mu}$ is highly unstable during training and does not affect the principal components of representations, so we remove it in our implementation.
The scaling term $(\sigma \cdot \gamma)^{-1}$ has a small magnitude and damages the optimization of training loss, so we re-normalize it to keep the sum of variances in transformed embeddings same as the original one.
We further introduce a hyper-parameter $\beta$, which controls the normalization strength, since it is shown in Section~\ref{sec:why_iso} that the dominating eigenvalue problem varies from datasets.
The modified IsoBN is written as:
\begin{align}
\label{eq:scaling1}
    \theta_i &= (\sigma_i \cdot \gamma_i + \epsilon)^{-\beta}, \\
    \label{eq:scaling2}
    \Bar{\theta} &= \frac{\sum_{i=1}^d \sigma_i^2}{\sum_{i=1}^d \sigma_i^2 \theta_i^2} \cdot \theta,\\
    \widehat{\bm{h}} &= \Bar{\theta} \odot \bm{h}.
\end{align}

\begin{table*}[!t]
\centering
\scalebox{0.9}{
\begin{tabular}{lccccccccc}
     \toprule
     Method & Avg& MNLI & QNLI &  RTE & SST-2 &
     MRPC & CoLA & STS-B \\
     \midrule \midrule
     BERT-base (ReImp) &81.37& 83.83 (.07) & 90.82 (.1)&  67.87 (1.1)& 92.43 (.7)&  85.29 (.9)& 60.72 (1.4)& 88.64 (.7) \\
     BERT-base-IsoBN  &\textbf{82.36}& \textbf{83.91} (.1) & \textbf{91.04} (.1)&   \textbf{70.75} (1.6)&\textbf{92.54} (.1)&  \textbf{87.50} (.6)& \textbf{61.59} (1.6)& \textbf{89.19} (.7) \\
     \midrule
     RoBERTa-L ({ReImp}) &88.16& 90.48 (.07) & 94.70 (.1) &   84.47 (1.0) & 96.33 (.3)& 90.68 (.9)& 68.25 (1.1)& 92.24 (.2) \\
     RoBERTa-L-IsoBN & \textbf{88.98} &\textbf{90.69 (.05)} & \textbf{94.91} (.1) &   \textbf{87.00} (1.3) & \textbf{96.67} (.3)& \textbf{91.42} (.8)& \textbf{69.70} (.8)& \textbf{92.51} (.2) \\
     \bottomrule
\end{tabular}}
\caption{\textbf{Empirical results on the dev sets of seven GLUE tasks.} We run 5 times with different random seeds and report median and std. 
IsoBN outperforms the conventional fine-tuning method around 1.0 absolute increment.}
\label{tab::main_results}
\end{table*}

In the IsoBN, the calculation of the scaling factor relies on the covariance and standard deviation statistics of the embedding.
We keep two moving average caches and update them in training because the estimated statistics from a single batch are not accurate.
The whole algorithm of IsoBN is shown in Algorithm~\ref{algo:isobn}.

\begin{algorithm}[!t]
\SetAlgoLined
\caption{IsoBN Transformation}
    \KwInput{Embedding $h$ over a mini-batch: $\mathcal{B}=\{h_{1...m}\}$; moving covariance $\Sigma$; moving standard deviation $\sigma$; momentum $\alpha$.}
    \KwOutput{transformed embedding $\widehat{h}$; updated $\Sigma$, $\sigma$.}
    \uIf{training}{
    $\mu_\mathcal{B} = \frac{1}{m}\sum_{i=1}^m h_i$\\
    $\sigma_\mathcal{B} = \sqrt{\frac{1}{m}\sum_{i=1}^m (h_i - \mu_\mathcal{B})^2}$ \\
    $\Sigma_\mathcal{B} = \frac{1}{m} \left(h - \mu_\mathcal{B}\right)^T \left(h - \mu_\mathcal{B}\right)$ \\
    $\sigma = \sigma + \alpha(\sigma_\mathcal{B} - \sigma)$\\
    $\Sigma = \Sigma + \alpha(\Sigma_\mathcal{B} - \Sigma)$ }
    $\rho = \frac{\Sigma}{\sigma \sigma^T}$ \\
    Compute $\gamma$ by Eq.~\ref{eq:gamma} \\
    Compute scaling factor $\theta$ by Eq.~\ref{eq:scaling1} and Eq.~\ref{eq:scaling2}\\
    $\widehat{h} = \theta \odot h$ \\
    \label{algo:isobn}
\end{algorithm}

\section{Evaluation}
\label{sec:exp}
In this section, we first present the setup of our experiments (i.e. the datasets, frameworks, and hyper-parameters), then discuss the empirical results, and finally evaluate the isotropy gain through the lens of explained variance. 
\subsection{Experiment Setup}
Our implementation of PTLMs is based on HuggingFace Transformer~\cite{Wolf2019HuggingFacesTS}.
The model is fine-tuned with AdamW~\cite{loshchilov2017decoupled} optimizer using a learning rate in the range of $\{1 \times 10^{-5}, 2 \times 10^{-5}, 5 \times 10^{-5}\}$ and batch size in $\{16, 32\}$.
The learning rate is scheduled by a linear warm-up~\cite{goyal2017accurate} for the first 6\% of steps followed by a linear decay to 0.
The maximum number of training epochs is set to 10.
For IsoBN, the momentum $\alpha$ is set to 0.95, the $\epsilon$ is set to 0.1, and the normalization strength $\beta$ is chosen in the range of $\{0.25, 0.5, 1\}$.

We apply early stopping according to task-specific metrics on the dev set.
We select the best combination of hyper-parameters on the dev set.
We fine-tune the PTLMs with 5 different random seeds and report the median and standard deviation of metrics on the dev set.

\subsection{Experimental Results}
We evaluate IsoBN on two PTLMs (BERT-base-cased and RoBERTa-large) and seven NLU tasks from the GLUE benckmark~\cite{wang2018glue}.
The experiments results are shown in Table~\ref{tab::main_results}.
Using IsoBN improves the evaluation metrics on all datasets.
The average score increases by 1\% for BERT-base and 0.8\% for RoBERTa-large.
For small datasets (MRPC, RTE, CoLA, and STS-B~\cite{Cer2017SemEval2017T1}), IsoBN obtains an average performance improvement of 1.6\% on BERT and 1.3\% on RoBERTa.
For large datasets (MNLI~\cite{Williams2018ABC}, QNLI~\cite{Rajpurkar2016SQuAD10}, and SST-2~\cite{Socher2013RecursiveDM}), IsoBN obtains an average performance improvement of 0.15\% on BERT and 0.25\% on RoBERTa.
This experiment shows that by improving the isotropy of embeddings, our IsoBN results in better fine-tuning performance.

\begin{table*}[!t]
\centering
\scalebox{1}{
\begin{tabular}{p{3.5cm}cccc}
     \toprule
     $EV_1 / EV_2 / EV_3 $ & MRPC& RTE& CoLA& STS-b \\
     \midrule \midrule
     BERT-base& 0.76 / 0.87 / 0.89& 0.88 / 0.93 / 0.95& 0.49 / 0.58 / 0.64& 0.89 / 0.94 / 0.96 \\
     BERT-base+BN& 0.74 / 0.84 / 0.86& 0.70 / 0.89 / 0.93& 0.37 / 0.59 / 0.63& 0.69 / 0.88 / 0.92 \\
     BERT-base+IsoBN& 0.37 / 0.68 / 0.77& 0.49 / 0.72 / 0.85& 0.25 / 0.37 / 0.48& 0.41 / 0.69 / 0.85 \\
     \midrule
     RoBERTa-L& 0.86 / 0.90 / 0.91& 0.53 / 0.66 / 0.70& 0.83 / 0.88 / 0.90& 0.87 / 0.90 / 0.92 \\
     RoBERTa-L+BN& 0.64 / 0.73 / 0.76& 0.36 / 0.50 / 0.57& 0.61 / 0.70 / 0.75& 0.65 / 0.72 / 0.77 \\
     RoBERTa-L+IsoBN& 0.18 / 0.36 / 0.43& 0.15 / 0.29 / 0.37& 0.21 / 0.38 / 0.49& 0.17 / 0.32 / 0.45 \\
     \bottomrule
\end{tabular}}
\caption{The explained variance on BERT-base and RoBERTa-large. Compared to batch normalization, our method can greatly reduce the explained variance and thus improve the isotropy of embeddings.}
\label{tab::EVK}
\end{table*}

\begin{figure*}[t]
  \centering
  \includegraphics[width=1\linewidth]{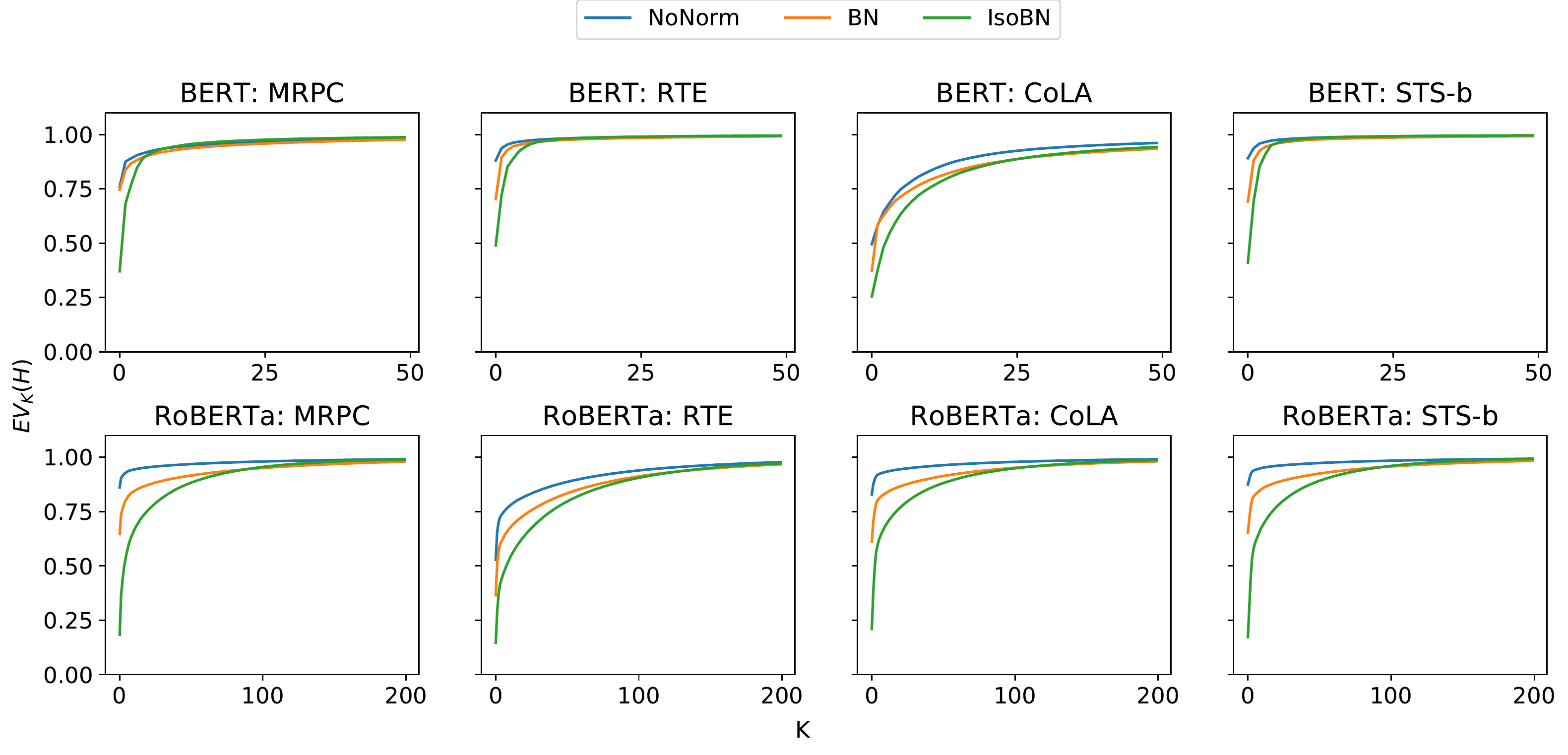}
  \caption{
  The $EV_k$ value on BERT-base and RoBERTa-large with no normalization, batch normalization, and IsoBN. We choose the max $K$ to be 50 for BERT and 200 for RoBERTa. Compared to batch normalization, IsoBN can greatly reduce the $EV_k$ value and thus improve the isotropy of \texttt{[CLS]} embeddings.}
  \label{fig::spectrum}
\end{figure*}

\subsection{Experiments of $EV_k$ Metric}

To quantitatively measure the isotropy of embeddings, we propose to use \textit{explained variance} (EV) as the metric for isotropy, which is defined as:
\begin{equation}
    EV_k(\bm{h}) = \frac{\sum_{i=1}^k \lambda_i^2}{\sum_{j=1}^d \lambda_j^2},
\end{equation}
where $\bm{h} \in \mathcal{R}^{N \times d}$ is the \texttt{[CLS]} embeddings, $\lambda_i$ is the $i^{th}$ largest singular value of the matrix $\bm{h}$.
Note that $N$ is the number of sentences in a certain corpus, and $d$ is the dimension of hidden states in the last layer of a pre-trained language model.

This metric measures the difference of variance in different directions of the embedding space.
Intuitively, if the $EV_k$ value is small, the variations of embedding tend to distribute equally in all directions and lead to more angular symmetric representation.
If the $EV_k$ value is large, most of the variations will concentrate on the first few directions, and the embedding space will degrade to a narrow cone.
Thus, the $EV_k$ is a good metric of the isotropy of embedding space.

We use $EV_k$ as the isotropy metric because it enjoys two beneficial properties:
\begin{itemize}
    \item It is invariant to the magnitude of the embeddings, and thus comparisons between different models and datasets is more fair.
    \item It is also invariant to the mean value of the embeddings, aligning with sentence classification/regression tasks of our interest.
\end{itemize}

We compute the $EV_k$ metric on two PTLMs (BERT-base and RoBERTa-large) and 4 tasks (MRPC, RTE, CoLA, STS-B). 
For IsoBN, the normalization strength $\beta$ is 1. 
We show the first three $EV_k$ value ($EV_1$, $EV_2$, and $EV_3$) in Table~\ref{tab::EVK}.

We observe that before normalization, the pre-trained \texttt{[CLS]} embeddings have very high $EV_k$ value.
The average $EV_3$ value is around 0.86 for both BERT-base and RoBERTa-large.
For some datasets (e.g. STS-b), the top three principal components already explain over 90\% of the variance.
Batch normalization can only reduce the $EV_k$ value by a small margin (0.025 for BERT and 0.148 for RoBERTa on $EV_3$), because it ignores the correlations among embedding dimensions.
Our proposed IsoBN greatly reduces the $EV_k$ value (0.123 for BERT and 0.425 for RoBERTa on $EV_3$).

We also visualize the distribution of $EV_k$ values in Figure~\ref{fig::spectrum}.
We choose the first 50 $EV_k$ value for BERT-base and first 200 $EV_k$ value for RoBERTa-large.
We observe that with IsoBN, we can decrease the $EV_k$ value of pre-trained embeddings.
This experiment shows that compared to batch normalization, IsoBN can further improve isotropy of \texttt{[CLS]} embedding of PTLMs.

\section{Related Work}
\subsection*{Input Normalization.}
Normalizing inputs~\cite{Montavon2012DeepBM,he2016deep,Szegedy2017Inceptionv4IA} and gradients~\cite{Schraudolph1998AcceleratedGD,Bjorck2018UnderstandingBN} has been known to be beneficial for training deep neural networks.
Batch normalization~\cite{ioffe2015batch} normalizes neural activations to zero-mean and unit-variance using batch statistics (mean, standard deviation).
It is both empirically and theoretically proved that batch normalization can greatly smooth the loss landscape~\cite{Santurkar2018HowDB,Bjorck2018UnderstandingBN,Ghorbani2019AnII}, which leads to faster convergence and better generalization.
One drawback of batch normalization is that it ignores the correlations between input dimensions.
Some methods~\cite{huang2018decorrelated,Huang2019IterativeNB} seek to calculate the full whitening transformation, while the IsoBN greatly simplifies the process by using the block-diagonal assumption, and succeeds to reduce the effect of dominating principal components by simple scaling of input dimensions.

\subsection*{Fine-tuning Language Models.}
Pre-trained language models~\cite{Devlin2019BERTPO,Liu2019RoBERTaAR} achieve the state-of-the-art performance on various natural language understanding tasks.
To adapt the pre-trained language model to target tasks, the common practice, in which a task-specific classifier is added to the network and jointly trained with PTLMs, already produces good results~\cite{Peters2019ToTO}.
Some works show that with well-designed fine-tuning strategies, the model performance can be further improved, by adversarial training~\cite{zhu2019freelb,Jiang2019SMARTRA}, gradual unfreezing~\cite{Howard2018UniversalLM,Peters2019ToTO}, or multi-tasking~\cite{Clark2019BAMBM,Liu2019MultiTaskDN}.
To our best knowledge, this is the first work to study the isotropy of text representation for fine-tuning.

\subsection*{Spectral Control.}
The spectrum of representation has been studied in many subareas.
On pre-trained word embeddings like Glove~\cite{Pennington2014GloveGV}, some work~\cite{mu2017allbutthetop} shows that removing the top one principal component leads to better performance on text similarity tasks.
On contextual embeddings like Elmo~\cite{Peters2018DeepCW} and GPT~\cite{Radford2018ImprovingLU}, spectrum is used as a measure of the ability to capture contextual knowledge~\cite{Ethayarajh2019HowCA}.
On text generation, it is shown that large eigenvalues hurt the expressiveness of representation and cause the degradation problem~\cite{gao2019representation,Wang2020ImprovingNL}.
In contrast, we study the spectrum of \texttt{[CLS]} embedding in pre-trained language models, and show that controlling the top principal components can improve fine-tuning performance.

\section{Conclusion}
Our major contributions in this paper are two-fold:
\begin{itemize} 
    \item We study the isotropy of the pre-trained \texttt{[CLS]} embeddings.
    Our analysis is based on straightforward visualization 
    about standard deviation and correlation coefficient.
    \item The proposed regularization method, IsoBN, stably improves the fine-tuning of BERT and RoBERTa towards more isotropic representations, yielding an absolute increment around 1.0 point on 7 popular NLU tasks. 
\end{itemize}
We hope our work points to interesting future research directions in improving pre-training language models as well as better fine-tuning towards more isotropy of PTLMs.
We will release our code at \textit{https://github.com/INK-USC/IsoBN/}.
\bibliography{aaai}
\end{document}